\documentclass{article}
\usepackage{spconf}%this is for ICIP/ICASSP template
\usepackage{amsmath,graphicx}
\usepackage{url,amsmath,amsfonts,amssymb,bm}

\usepackage{enumitem}
\usepackage{graphicx}
\usepackage{algorithm}
\usepackage{algorithmicx, algpseudocode}
\usepackage{multirow}
\usepackage{color}
\usepackage{mathtools}

\usepackage{caption}
\usepackage{subcaption}

\usepackage{amsthm}
\newtheorem{definition}{Definition}

\newtheorem{theorem}{Theorem}
\newtheorem{lemma}{Lemma}

\newtheorem{proposition}{Proposition}

 \DeclareMathOperator*{\erank}{{r}}

\DeclareMathOperator*{\tr}{tr}
\DeclareMathOperator*{\diag}{diag}

\DeclareMathOperator*{\E}{\mathbb{E}}
\DeclareMathOperator*{\Cov}{Cov}
\DeclareMathOperator*{\Prob}{\mathbb{P}}

\title{Active covariance  estimation by random  sub-sampling of variables}
\name{Eduardo Pavez and Antonio Ortega
\address{University of Southern California}
\thanks{This work was supported in part by NSF under grant CCF-1410009. Contact author e-mail: pavezcar@usc.edu, ortega@sipi.usc.edu.}
}
\begin{document}
\ninept
\maketitle
\begin{abstract}
We study covariance matrix estimation for the case of partially observed random vectors, where different samples contain different subsets of vector coordinates.  Each observation is  the product of the variable of interest with  a $0-1$ Bernoulli random variable.  We analyze an unbiased covariance estimator under this model,  and derive an error bound that reveals relations between the sub-sampling probabilities and the entries of the covariance matrix.  We apply our analysis in an active learning framework, where the expected number of observed variables   is small compared to the dimension of the vector of interest, and propose a design of  optimal  sub-sampling probabilities and an active covariance matrix estimation algorithm. 
\end{abstract}
\section{Introduction}
\label{sec:intro}
 We study  estimation of the covariance matrix   of a random vector $\mathbf{x}$ from observations of the form $\mathbf{y} = [\delta_1 x_1, \delta_2 x_2, \cdots, \delta_n x_n]^{\top}$,
where each component $x_i$ is multiplied by a  $0-1$ Bernoulli random variable $\delta_i$. We assume the probabilities $\Prob(\delta_i=1)=p_i$ are  known, independent of each other, and of $\mathbf{x}$. Partial observations consistent with this  model may arise when there are some physical limitations to the observation process (and thus there is missing data) or when the cost of observation needs to be reduced by selecting what to sample (i.e., active learning).  

 Applications where probabilistic models need to be constructed from observations with missing data include transportation networks \cite{asif2016matrix} and  sensor networks \cite{deshpande2004model}. For example, different sensors in the network might have have different capabilities, e.g. different reliability or  number of samples per hour that can be acquired, and these can be captured by associating different sub-sampling probabilities $p_i$ to each sensor. 

%Active learning  usually refers  to semi supervised classification problems, 
For covariance estimation, or more generally  graphical model estimation, active learning approaches have been considered where variables are observed in a sequential and adaptive manner to identify the  graphical model  with minimal number of observations.  Active learning approaches for covariance estimation can be useful in distributed computation environments like sensor networks,  where there are acquisition, processing or communication constraints, and there is a need for optimal resource allocation. 

One critical difference between missing data and active covariance estimation  scenarios is the control over the observation model.  In the missing data case,  the sub-sampling probabilities are a problem feature that is given or has to be estimated from data, while in active learning problems, the sampling probabilities are a design parameter. The results from this paper can be useful for analysis in the missing data case, as well as for designing active learning algorithms as we will discuss in more detail in Section \ref{sec_adaptive}.

In this work we analyze an unbiased covariance matrix estimator under sub-Gaussian assumptions on $\mathbf{x}$. Our main result is an  error bound on the Frobenius norm that reveals the relation between number of observations, sub-sampling probabilities and entries of the true covariance matrix.  We apply this error bound to the design of sub-sampling probabilities in an active covariance estimation scenario.  An interesting conclusion from this work is that when the covariance matrix is approximately low rank, an active covariance estimation approach can perform almost as well as an estimator with complete observations.  The paper is organized as follows, in  Section \ref{sec_related} we review related work. Sections \ref{sec_estimator} and \ref{sec_concentration} introduce the problem and the main results respectively. In Section \ref{sec_adaptive} we show an application of our bounds for  optimal design of sampling probabilities for a batch based active covariance estimation algorithm. Proofs are presented in Section \ref{sec_proofs} and  conclusion in Sections \ref{sec_conclusion}.
\section{Related work}
\label{sec_related}
 Lounici \cite{lounici2014high} studies  covariance matrix estimation  from missing data when all variables can be observed with the same probability. The focus of \cite{lounici2014high} is on estimation of approximately low rank covariance matrices with a matrix version of the lasso algorithm. In \cite{jurczak2017spectral} the  empirical covariance matrix under missing data is shown to be indefinite. Most recently Cai and Zhang \cite{cai2016minimax} study the  missing completely at random model (MCAR), which assumes arbitrary and unknown missing data probabilities. They show that a modified sample covariance matrix estimator, similar to ours but using the empirical estimation of probabilities $p_i$, achieves optimal minimax rates in spectral norm for bandable and sparse covariance matrices. 
  Compared to \cite{lounici2014high,cai2016minimax,jurczak2017spectral},  we allow all probabilities to be different and known, and we only study the unbiased sample covariance estimator. Moreover, all of the aforementioned papers consider estimation errors in spectral norm, while we consider   errors in Frobenius norm,  allowing us to derive  error bounds with precise dependences between the entries of the true covariance matrix and the sampling probabilities $p_i$. These bounds can then be applied to design the sampling distribution in an adaptive manner.
 
For distributed computing applications, the work of \cite{azizyan2015extreme}  studies covariance estimation from random subspace projections using dense matrices, which are generalized in \cite{pourkamali2016estimation} to sparse projection matrices. The same approach from \cite{pourkamali2016estimation} is used in \cite{anaraki2014memory} for memory and complexity reduced PCA. In all these works \cite{azizyan2015extreme,anaraki2014memory, pourkamali2016estimation},  each  measurement is a (possibly sparse) linear combination of a few variables, and even though the designs are random, they have a  fixed distribution for all observations.   

The work of \cite{kolar2012consistent} considers the case when all $p_i$ are unknown, and uses an unbiased covariance matrix estimator  as an input to the graphical lasso algorithm \cite{friedman_sparse_2008} for inverse covariance matrix estimation. Other interesting  active learning approaches for graphical model selection are \cite{dasarathy2016active},   \cite{vats2014active} and \cite{scarlett2017lower}.  Vats et.~al.~\cite{vats2014active} uses a method that combines sampling marginals with conditional independence testing to learn the graphical model structure. \cite{dasarathy2016active,scarlett2017lower} consider a more general family of algorithms based on sampling high degree vertices, and prove upper and lower performance bounds. 
% * <antonio.ortega@sipi.usc.edu> 2017-10-28T07:16:16.839Z:
% 
% The last sentence of the above paragraph is vague and the overall sense after reading the paragraph is that it's not so clear how different our work is, while before it was clearly stated that other methods did linear combinations, and we don't. Can we say that many of these "interesting approaches" do this? 
% 
% ^.
 
 Random sub-sampling and reconstruction of signals has been studied within graph signal processing \cite{chen2016signal, puy2016random} and statistics \cite{loh2012high}. The Bernoulli observation model we use has been studied by \cite{chen2016signal, puy2016random} as a sampling strategy for  graph signals, where sampling probability designs are proposed for  reconstruction of deterministic band-limited signals. Also,  Romero et.~al.~\cite{romero2016compressive} derives covariance matrix estimators  assuming the target covariance matrix is a linear combination of known covariance matrices, which leads to algorithms and theoretical analysis  that 
 are fundamentally different from ours. 
%
%\subsection{Signal reconstruction by interpolation/imputation}
%The probailistic model of data acquisition has been considered in the graph signal procesing community in the context of sampling and interpolation of signals on graphs.  Particular interest are the Siheng paper \cite{chen2016signal}, where uses random sampling of bandlimited deterministic graph signals, they also design sampling probabilities in a MMSE sense. This paper is also in the same spirit, \cite{di2016adaptive}.
%From statistics, we have the Loh paper \cite{loh2012high} for signal reconstruction (sort of a generalized Lasso).
%
%Also review papers in data imputation/completion.
\section{Problem Formulation}
\label{sec_estimator}
\subsection{Notation}
We denote scalars using regular font, while we use bold for vectors and matrices, e.g.,  $\mathbf{a}=(a_i)$, and $\mathbf{A}=(a_{ij})$. The Hadamard product between matrices is defined as $(\mathbf{A} \odot \mathbf{B})_{ij} = a_{ij} b_{ij}$. We use $\Vert \cdot \Vert_q$  for entry-wise matrix norms, with  $q=2$ corresponding to  the Frobenius norm. $\Vert \cdot \Vert$ denotes  $\ell_2$ norm or spectral norm when applied to vectors or matrices respectively. 
\subsection{Unbiased  estimation}
Consider  a random vector $\mathbf{x}$ taking values in $\mathbb{R}^n$. We observe 
\begin{align}\label{eq_observation}
\mathbf{y} = \boldsymbol{\delta} \odot \mathbf{x},
\end{align}
where $ \boldsymbol\delta = (\delta_i)$ is a vector of Bernoulli $0-1$ random variables. The probability of observing the $i$-th variable is given by $\Prob(\delta_i =1)=p_i$. The vector of probabilities is denoted by $\mathbf{p} = [p_1,\cdots, p_n]^{\top}$, and $\mathbf{P} = \diag(\mathbf{p})$ is a diagonal matrix.
%We say  $\mathcal{S}= \lbrace i: \delta_i \neq 0 \rbrace$ is the sampling set and $\mathcal{S}^c = \mathcal{I}$ is the interpolation set. 
The average number of samples corresponds to $\E(\sum_{i=1}^n \delta_i)=\sum_{i=1}^n p_i = m$.
Given  $\mathbf{y}^{(1)},\cdots, \mathbf{y}^{(T)}$, i.i.d. realizations of $\mathbf{y}$, the $i$-th variable will be sampled in average $p_i T$ times, if all $p_i = 1$, then $\mathbf{y=x}$, and we have perfect observation of $\mathbf{x}$.  We are interested in studying covariance estimation for $\mathbf{x}$ when   $m <n$ and  $0<p_i$ for all $i$.
 Let $\boldsymbol \mu$ and $\mathbf{\Sigma}$ be the mean and covariance of $\mathbf{x}$, then
\begin{align*}
\E(\mathbf{y}) = \mathbf{P} \boldsymbol\mu, \textnormal{ and }
\Cov(\mathbf{y}) = \mathbf{\Sigma} \odot \mathbf{\Xi} + (\mathbf{P-P}^2)\diag(\boldsymbol\mu \boldsymbol\mu^{\top}),
\end{align*}
where $\mathbf{\Xi} = (\xi_{ij})$ is  defined as $\xi_{ii} = p_i$ and $\xi_{ij} = p_i p_j$ when $i \neq j$.
%\[  \xi_{ij}= \left\{
%\begin{array}{ll}
%      p_i &  i = j \\
%	       p_i p_j & i \neq j, \\
%	\end{array} 
%	\right. \]

%
%By  decomposing as  $\mathbf{\Xi} = \mathbf{p p}^{\top} -\mathbf{P}^2+\mathbf{P} $, it is easy to see that $\mathbf{\Xi} \succ 0$, however its Hadamard (entrywise) inverse   $\mathbf{\Xi}^{\dagger}$ is negative indefinite.
%
%
For the rest of the paper we will assume $\boldsymbol\mu = \mathbf{0}$.
Given a set of  of i.i.d. samples $\lbrace \mathbf{y}^{(k)} \rbrace_{k=1}^T$ of the random vector $\mathbf{y}$,  define
\begin{align}\label{eq_est_cov_y_mu}
 \widehat{\mathbf{\Sigma}} = \frac{1}{T} \sum_{k=1}^T \mathbf{y}^{(k)} {\mathbf{y}^{(k)}}^{\top}  \odot \mathbf{\Xi}^{\dagger},
\end{align}
where $\mathbf{\Xi}^{\dagger}$ is the Hadamard (entry-wise) inverse of $\mathbf{\Xi}$. A simple calculation shows that $\widehat{\mathbf{\Sigma}}$ is an unbiased estimator for $\mathbf{\Sigma}$. Indeed, $\E(\widehat{\mathbf{\Sigma}}) = \frac{1}{T} \sum_{k=1}^T \E(\mathbf{y}^{(k)} {\mathbf{y}^{(k)}}^{\top} ) \odot \mathbf{\Xi}^{\dagger} = \mathbf{\Sigma} \odot \mathbf{\Xi}  \odot \mathbf{\Xi}^{\dagger} = \mathbf{\Sigma} $.
Because $\mathbf{ \Xi}^{\dagger} \nsucceq 0$, the matrix $\widehat{\mathbf{\Sigma}}$ might not be positive semi-definite (conditions for $\widehat{\mathbf{\Sigma}}$ to be positive semi-definite are given in \cite{jurczak2017spectral}). 
\section{Estimation error}
\label{sec_concentration}
In this section  we present an  error analysis of the covariance matrix estimator from (\ref{eq_est_cov_y_mu})  when  $\mathbf{x}$ has sub-Gaussian entries. Sub-Gaussian random variables include Gaussian, Bernoulli and Bounded random variables.  For more information  see \cite{vershynin2016high} and references therein.  
%
%\subsection{sub-Gaussian and sub-exponential random variables}
\begin{definition}[\cite{vershynin2016high}]
If $\E[\exp(z^{2} /K^{2})] \leq 2$ holds for  some $K>0$, we say $z$ is sub-Gaussian. If  $\E[\exp(\vert z \vert /K)] \leq 2$ holds for some $K>0$, we say $z$ is sub-exponential.
\end{definition} 
\begin{definition}[\cite{vershynin2016high}]
The sub-Gaussian and sub-exponential norms are defined as
\begin{align*}
\Vert z \Vert_{\psi_{\alpha}} = \inf\lbrace u>0 : \E[\exp(\vert z\vert^{\alpha}/u^{\alpha})] \leq 2
 \rbrace.
\end{align*}
for $\alpha =2$ and $\alpha =1$ respectively.
\end{definition}
%
% * <antonio.ortega@sipi.usc.edu> 2017-10-28T07:25:22.156Z:
% 
% I'm not familar with the \psi_\alpha notation. Can we say something about it? Or is this just a standard notation for these norms?
% 
% ^ <eduardo.pavez.carvelli@gmail.com> 2017-10-28T08:20:28.758Z:
% 
% this is standard, i follow the same convention as the book "High dimensional probability"
%
% ^.
Sub-Gaussian and sub-exponential random variables, and their norms,  are related as follows.
\begin{proposition}[\cite{vershynin2016high}]
\label{prop_subgauss_subexp}
If $z$ and $w$ are sub-Gaussian, then $z^2$ and $zw$ are sub-exponential with norms satisfying $\Vert z^2 \Vert_{\psi_1} = \Vert z \Vert_{\psi_2}^2$, and $\Vert zw \Vert_{\psi_1} \leq \Vert z \Vert_{\psi_2} \Vert w \Vert_{\psi_2}$.
\end{proposition}
We have that the following characterization of the product of  sub-Gaussian and Bernoulli random variables.
\begin{lemma}\label{lemma_bern_subgauss}
Let $y_1 = \delta_1 x_1$ and $y_2 = \delta_2 x_2$ be a product of Bernoulli $\delta_1, \delta_2$ and sub-Gaussian $x_1, x_2$ random variables with Bernoulli probabilities $p_1$ and $p_2$ respectively, the only dependent variables are $x_1$ and $x_2$, then
\begin{enumerate}
\item[1)]    $y_i$ is sub-Gaussian and $\Vert y_i \Vert_{\psi_2} 	\leq \Vert x_i \Vert_{\psi_2}$.
\item[2)]  $y_1^2$, $y_2^2$, and $y_1 y_2$ are sub-exponential with norms satisfying  $\Vert y_i y_j \Vert_{\psi_1} \leq \Vert x_i x_j \Vert_{\psi_1}$ for $i,j =1,2$.
\end{enumerate}
%
%\begin{proof}
%To show 1), we start with the following inequality for  $y_i$
%\begin{align} \nonumber
%\E[\exp(  y_i^2 /\Vert x_i \Vert_{\psi_2}^2)] &= p_i\E[\exp(x_i^2/\Vert x_i \Vert_{\psi_2}^2)] + (1-p_i) \\
%&\leq 2p_i +(1-p_i)= p_i+1 \leq 2.\label{eq_bern_gauss_prod}
%\end{align}
%Where the first inequality holds because $x_i$ is sub-Gaussian. The inequality $\Vert y_i \Vert_{\psi_2} 	\leq \Vert x_i %\Vert_{\psi_2}$ follows from the definition of sub-Gaussian norm and (\ref{eq_bern_gauss_prod}).
%See Section \ref{app_proof_lemma_bern_subgauss}
%The proof of 2) follows similar steps using the definition of sub-exponential norm, and the fact that $x_i x_j$ are sub-%exponential.
%\end{proof}
\end{lemma}
The proof of Lemma \ref{lemma_bern_subgauss} can be easily obtained from the definition of sub-Gaussian and sub-exponential norms. We omit it for space considerations.
%We will need the  quantity $\kappa_{ij} = \log(\E[\exp(\vert x_i x_j\vert/\Vert x_i x_j \Vert_{\psi_1}].$
%
%
%
%
%
%\subsection{Consistency in entry-wise $\ell_{q}$ norm}
%
We also  define the matrix  $\mathbf{H} = (h_{ij})$, with entries given by
$$
h_{ij} = \left\{
        \begin{array}{ll}
            \frac{\Vert x_i x_j \Vert_{\psi_1}}{p_i p_j} & \quad i \neq j \\
            \frac{\Vert x_i^2 \Vert_{\psi_1}}{p_i} & \quad  i=j.
        \end{array}
    \right.
$$
Now we state our main result, whose proof appears in Section \ref{sec_proofs}.
\begin{theorem}\label{th_concistency_fro}
Let $\mathbf{x}$ be  zero mean random vector in $\mathbb{R}^n$ with sub-Gaussian entries and norm $\Vert x_i \Vert_{\psi_2}$. Let $\mathbf{y } = \boldsymbol\delta \odot \mathbf{x}$, where each $\delta_i$ is a Bernoulli random variable with parameter $0<p_i\leq 1$, independent of each other and of $\mathbf{x}$.   Given i.i.d. realizations $\lbrace \mathbf{y}^{(k)} \rbrace_{k=1}^T $, the estimator from (\ref{eq_est_cov_y_mu}) satisfies
\begin{align*}
\Vert \hat{\mathbf{\Sigma}} - \mathbf{\Sigma}\Vert_q \leq \Vert \mathbf{H} \Vert_q \left\lbrace  \sqrt{\gamma\frac{2\log(n)+\log(\eta)}{T}}  \vee  \gamma \frac{2\log(n)+\log(\eta)}{T}   \right\rbrace
\end{align*}
  with probability at least $1-\frac{2}{\eta}$, where $\gamma$ is an universal constant. Moreover if $\Vert x_i \Vert_{\psi_2} = \sigma \sqrt{\Sigma_{ii}}$ and $q \geq 2$, then
\begin{align}\label{eq_theo_effRank_bound}
\Vert \mathbf{H} \Vert_q \leq \frac{2\sigma^2}{\hat{p}^2} \erank(\mathbf{\Sigma}) \Vert \mathbf{\Sigma}\Vert,
\end{align}
where $\hat{p} = \min{p_i}$, and $\erank(\mathbf{\Sigma}) = \tr(\mathbf{\Sigma})/\Vert \mathbf{\Sigma}\Vert$ is the {\bf effective rank}.
%\begin{proof}
%See Section \ref{proof_consistency_fro}.
%\end{proof}
\end{theorem}
Theorem \ref{th_concistency_fro} shows that the estimation error $\Vert\hat{\mathbf{\Sigma}} - \mathbf{\Sigma}\Vert_q \rightarrow 0$ in probability as the number of samples increases. More importantly, our result reveals that the  sampling probabilities $p_i$ are closely related to the sub-exponential norms of the variables $x_i x_j$ through the matrix $\Vert  \mathbf{H} \Vert_q$. The   bound from (\ref{eq_theo_effRank_bound} )  suggests that distributions with smaller effective rank \cite{lounici2014high,vershynin2016high} can tolerate a more aggressive sub-sampling factor (smaller $m$).  This is not surprising since the  effective rank  $\erank(\mathbf{\Sigma})$  is upper bounded by the actual rank, and can be significantly smaller for distributions whose energy concentrates in  few principal components. We also note that the ratio $\erank(\mathbf{\Sigma})/\hat{p}^2$  appears in the bounds of \cite{lounici2014high} as well.
%\begin{proposition}\label{prop_bound_H}
%If  $\Vert x_i \Vert_{\psi_2} = \sigma \sqrt{\Sigma_{ii}}$ and $q \geq 2$. Then $\Vert \mathbf{H} \Vert_q$ from Theorem \ref{th_concistency_fro} satisfies
%\begin{align*}
%\Vert \mathbf{H} \Vert_q \leq \frac{2\sigma^2}{\hat{p}^2} \erank(\mathbf{\Sigma}) \Vert \mathbf{\Sigma}\Vert,
%\end{align*}
%where $\hat{p} = \min{p_i}$, and $\erank(\mathbf{\Sigma}) = \tr(\mathbf{\Sigma})/\Vert \mathbf{\Sigma}\Vert$ is the {\bf effective rank}.
%%
%\end{proposition}
%
%Theorem \ref{th_concistency_fro} shows that the sampling probabilities $p_i$ are closely related to the sub-exponential norms of the variables $x_i x_j$. The more crude bound from Proposition \ref{prop_bound_H} suggests that distributions with smaller effective rank \cite{lounici2014high,vershynin2016high} can tolerate a more aggressive sub-sampling factor. This is not surprising since the  effective rank  $\erank(\mathbf{\Sigma})$  is upper bounded by the actual rank, and can be significantly smaller for distributions whose energy concentrates in  few principal components. We also note that the ratio $\erank(\mathbf{\Sigma})/\hat{p}^2$  appears in the bounds of \cite{lounici2014high} as well.
%
%
\begin{algorithm}[t]
\caption{ Covariance estimation with adaptive sampling}\label{alg1}
\begin{algorithmic}[1]
\Require initial distribution $ \mathbf{p}^{(0)}$, covariance $\widehat{\mathbf{\Sigma}}^{(0)} = \mathbf{0}$,  budget $\mathbf{1}^{\top}\mathbf{p}^{(0)} =m$, and batch size $B$.
%\PROCEDURE{GLP estimation}{}
%\While{not converged}
\For{$t=0$ to $N-1$}
\State Sample $B$ i.i.d. realizations of $\mathbf{y} = \boldsymbol\delta(\mathbf{p}^{(t)}) \odot \mathbf{x}$
\State Estimate $\widehat{\mathbf{\Sigma}}$
\State Update $\widehat{\mathbf{\Sigma}}^{(t+1)} \gets \frac{1}{t+1} \widehat{\mathbf{\Sigma}} + \frac{t}{t+1}\widehat{\mathbf{\Sigma}}^{(t)}$
\State Update $\mathbf{p}^{(t+1)}$ by solving (\ref{eq_opt_p_2}) with $\widehat{\mathbf{\Sigma}}^{(t+1)}$ and budget $m$.
\EndFor
%\EndWhile 
%\EndProcedure
\end{algorithmic}
\end{algorithm}
\section{active covariance estimation}
\label{sec_adaptive}
In this section we consider a scenario where we cannot observe (on average) more than $m$ variables at a time, but we have the freedom to choose the sub-sampling probability distribution.
\subsection{Sub-sampling distribution  for true covariance matrix}
Based on the error bound from Theorem \ref{th_concistency_fro}, we propose designing the sub-sampling distribution by approximately  minimizing $\Vert \mathbf{H}\Vert_q$.  Theorem \ref{th_concistency_fro} suggests that  $p_i$ should be larger whenever the sub-Gaussian norm of $x_i$ is large, but also the product $p_i p_j$ should be large when the sub-exponential norm of $x_ix_j$ is large. We assume that $\Vert x_i \Vert_{\psi_2} = \sigma \sqrt{\Sigma_{ii}}$. The  bounds for $h_{ii}$ and $h_{ij}$  from (\ref{eq_bound_hii}) and (\ref{eq_bound_hij}) respectively suggest  the approximation $p_i^{2} \sim \Sigma_{ii}$. Given a sampling budget $m$, we estimate the sub-sampling probability vector $\mathbf{p}$ by solving the following scaled projection problem
\begin{align}\label{eq_opt_p}
\min_{\mathbf{p}, \rho} \frac{1}{2}\Vert \mathbf{p} - \rho \diag(\mathbf{\Sigma})^{\frac{1}{2}} \Vert_2^2, \textnormal{ s.t. } &\mathbf{1}^{\top}\mathbf{p}=m,  \mathbf{0 \leq p \leq 1}.
\end{align}
% * <antonio.ortega@sipi.usc.edu> 2017-10-28T07:32:02.369Z:
%
% ^.
%
\subsection{Sub-sampling distribution for empirical covariance matrix}
Since the true covariance matrix is unknown, (\ref{eq_opt_p}) does not lead to a practical estimator. Instead, we  consider a batch based algorithm that for a given   budget $m$, and a   starting sub-sampling distribution, it iteratively refines the sub-sampling probability distribution as a function of previous observations. We show the pseudo code for such procedure in Algorithm \ref{alg1}, which can be summarized in  the following steps: observation with variable sub-sampling, covariance estimation, and sub-sampling distribution update. At the $t$-th iteration, $B$ i.i.d. realizations are observed according to (\ref{eq_observation}) with sub-sampling probabilities $\mathbf{p}^{(t)}$. The covariance estimator is a convex combination of the estimator at the previous iteration, and the estimator for the  current batch.  %Note that any convex combination of unbiased estimators is also unbiased. 
Finally,  the new covariance matrix estimator is used to update the sub-sampling probabilities as 
\begin{align}\label{eq_opt_p_2}
\mathbf{p}^{(t)} = &\arg\min_{\mathbf{p}, \rho} \frac{1}{2}\Vert \mathbf{p} - \rho \diag(\widehat{\mathbf{\Sigma}}^{(t)})^{\frac{1}{2}} \Vert_2^2, \\
&\textnormal{ s.t. } \mathbf{1}^{\top}\mathbf{p}=m,  \mathbf{0 \leq p \leq 1}. \nonumber
\end{align}
\begin{proposition} 
 Algorithm \ref{alg1} produces an unbiased estimator for $\mathbf{\Sigma}$.
\begin{proof}
We will proceed by induction. For the first iteration we have used uniform sampling, thus we have that $\mathbb{E}(\widehat{\mathbf{\Sigma}}^{(1)}) = \mathbf{\Sigma}$. Now assume $\mathbb{E}(\widehat{\mathbf{\Sigma}}^{(t)}) = \mathbf{\Sigma}$, then at the $(t+1)$-th iteration, the covariance of the new data satisfies $\mathbb{E}(\widehat{\mathbf{\Sigma}})=\mathbf{\Sigma}$, which implies $\mathbb{E}(\widehat{\mathbf{\Sigma}}^{(t+1)}) =\frac{1}{t+1}\mathbb{E}(\widehat{\mathbf{\Sigma}})+\frac{t}{t+1}\mathbb{E}(\widehat{\mathbf{\Sigma}}^{(t)}) = \mathbf{\Sigma} $.
\end{proof}
\end{proposition}
\subsection{Numerical evaluation}
In this section we evaluate our proposed method using the MNIST \cite{lecun1998gradient} dataset, which consists of $28 \times 28$ images of scanned digits from $0$ to $9$. In our experiments we consider $N=5851$ images of  the digit $8$. The  vectorized, and mean  removed images are denoted by $\lbrace\mathbf{z}_i\rbrace_{i=1}^{N}$ with covariance matrix $\mathbf{C}$.  We consider estimation of the  the covariance matrix of $\mathbf{x}_i = \mathbf{z}_i + \sqrt{\theta \Vert \mathbf{C} \Vert} \mathbf{e}$, where $\mathbf{e}$ is zero mean Gaussian noise with unit variance,  therefore $\mathbf{\Sigma} = \mathbf{C} + \theta \Vert \mathbf{C} \Vert \mathbf{I}$.

\begin{figure*}[t]
\centering
\begin{subfigure}[b]{.33\textwidth}
\includegraphics[width = \textwidth]{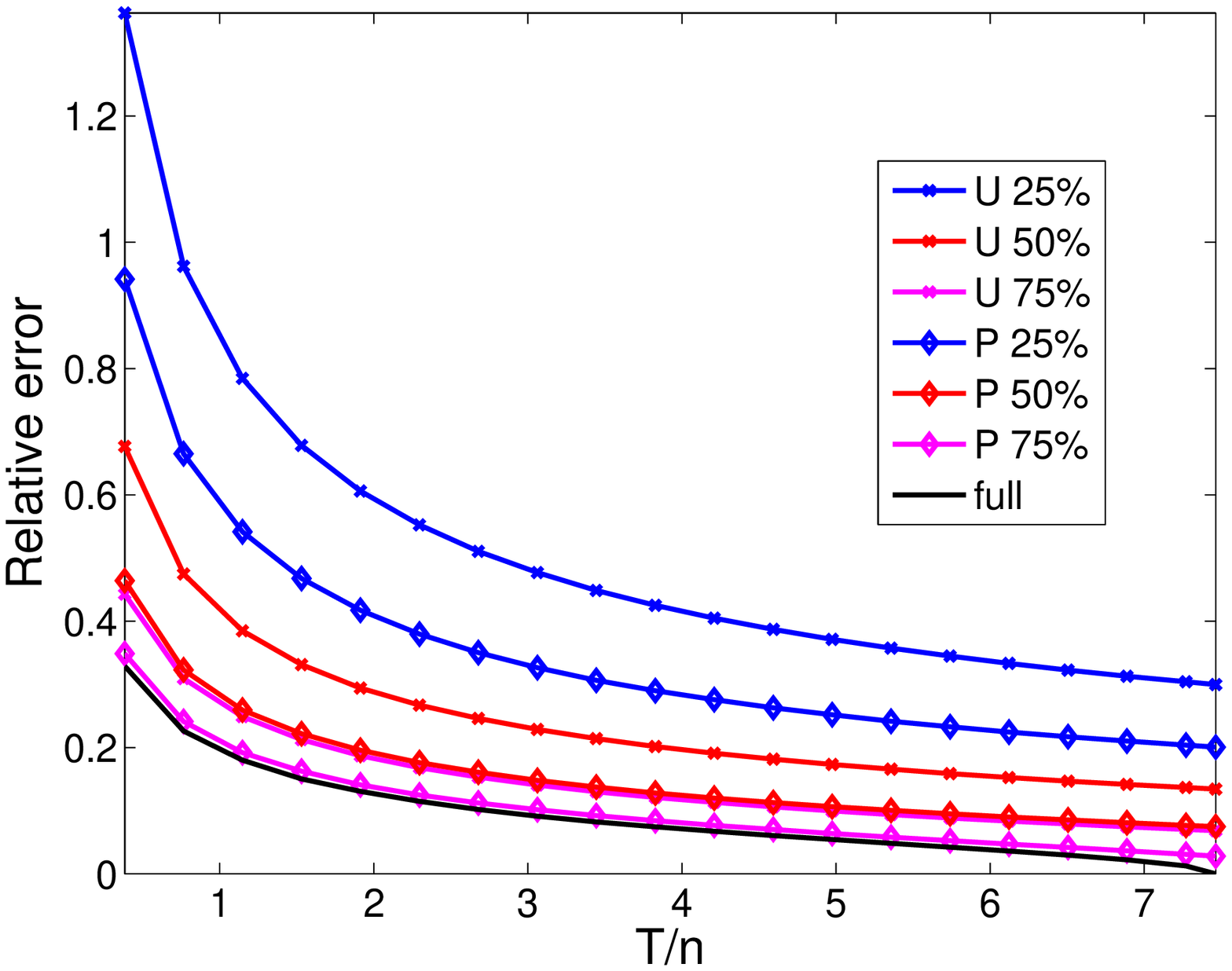}
\caption{Uniform vs  non-uniform sub-sampling}
\label{fig_fixed}
\end{subfigure}
\begin{subfigure}[b]{.33\textwidth}
\includegraphics[width = \textwidth]{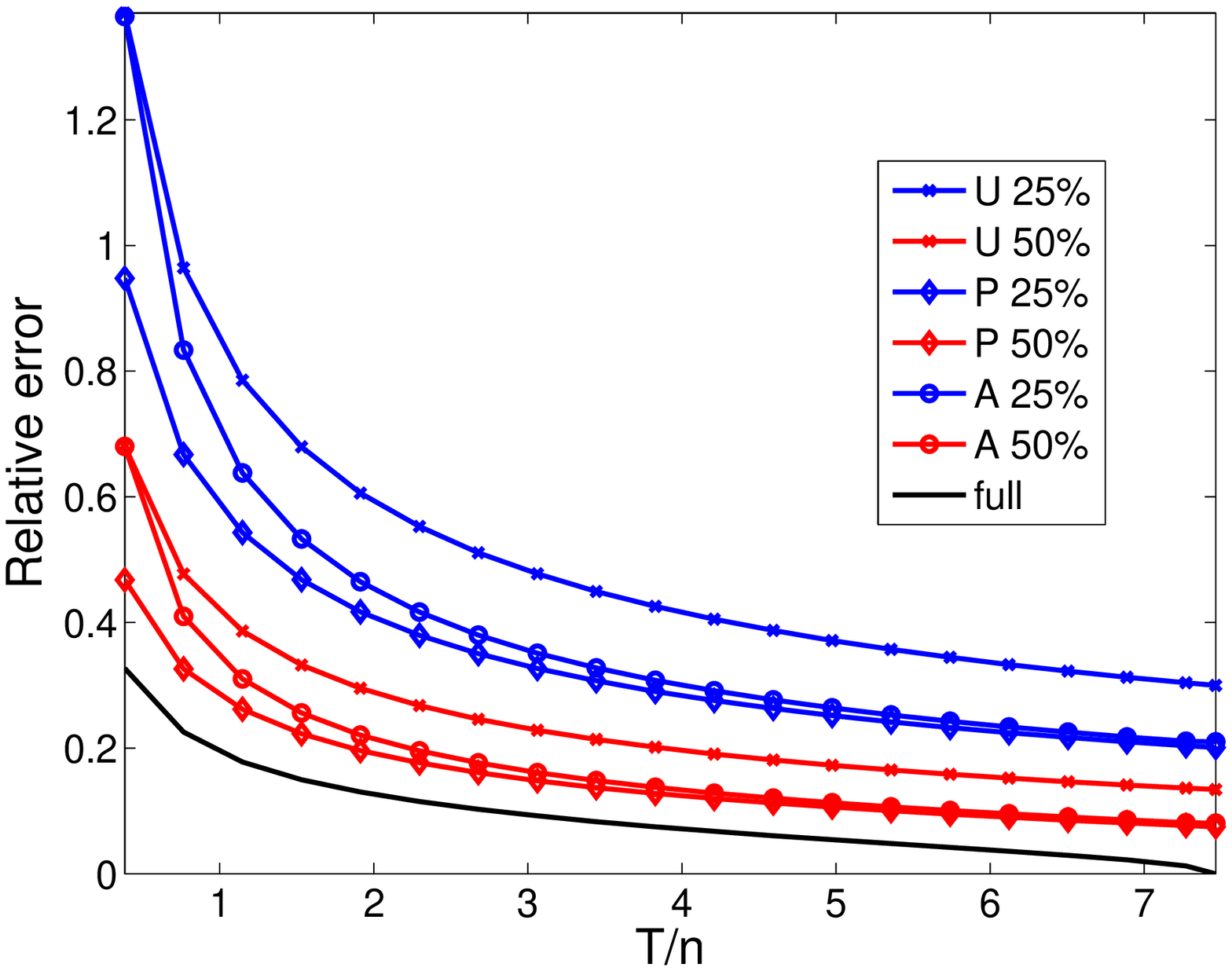}
\caption{Active covariance estimation, $\theta =1/n$}
\label{fig_adaptive}
\end{subfigure}
\begin{subfigure}[b]{.33\textwidth}
\includegraphics[width = \textwidth]{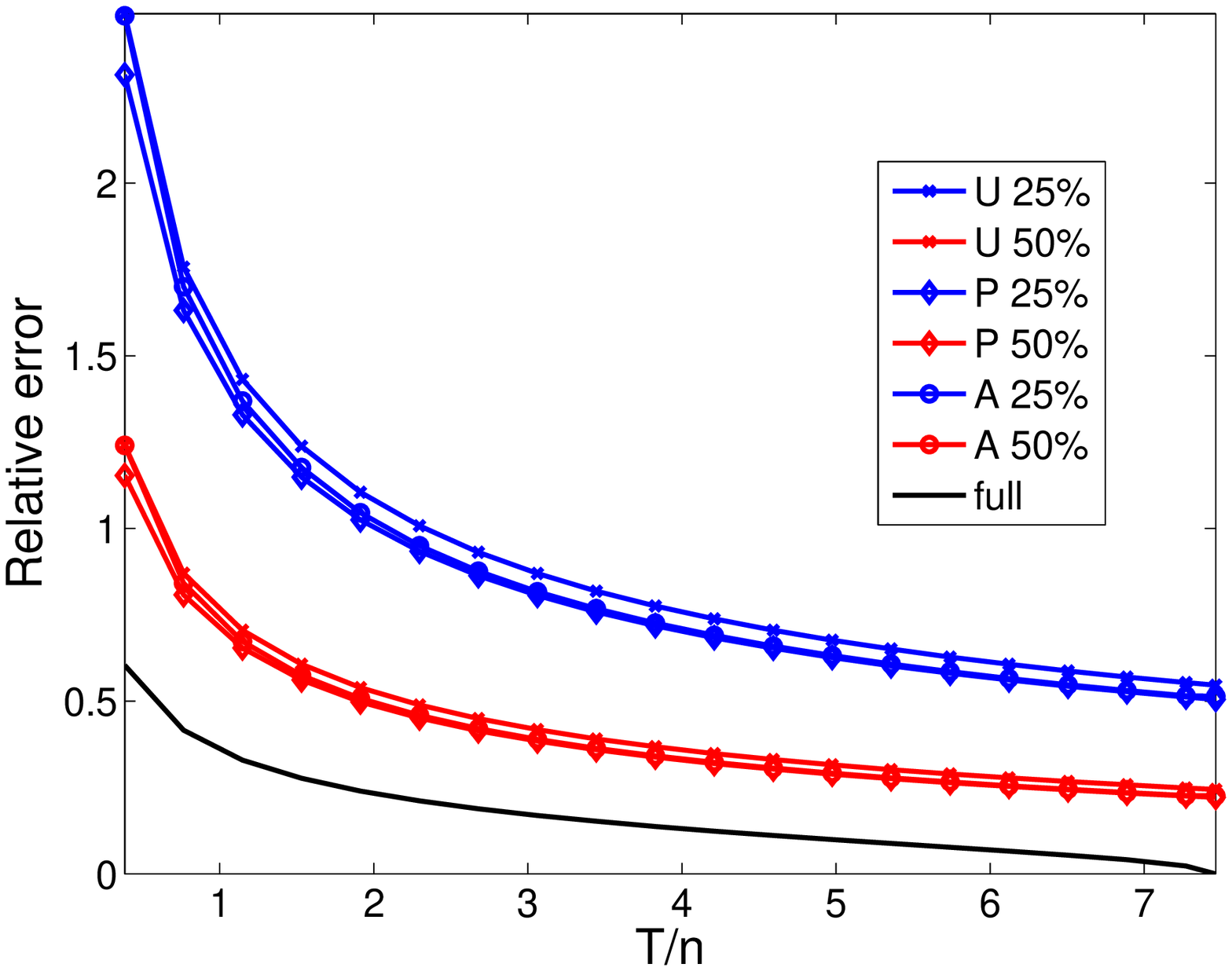}
\caption{Active covariance estimation, $\theta =10/n$}
\label{fig_adaptive2}
\end{subfigure}
\caption{Performance of different covariance estimators in relative Frobenius norm. (U) corresponds to uniform sub-sampling $p_i = m/n$, (P) denotes non-uniform sub-sampling found using (\ref{eq_opt_p}), (A) denotes active covariance estimation using Algorithm \ref{alg1}}.
\end{figure*}

To draw i.i.d. realizations of $\mathbf{x}$, we sample images $\mathbf{z}_i$ without replacement and add Gaussian noise with variance $\theta \Vert \mathbf{C} \Vert$.
The effective rank of $\mathbf{\Sigma}$ is controlled by the parameter $\theta$ and satisfies
\begin{align*}
\erank(\mathbf{\Sigma}) = \frac{\erank( \mathbf{C})+ n\theta}{1 + \theta}.
\end{align*}
We first compare uniform sub-sampling with non uniform sub-sampling for estimation of $\mathbf{\Sigma}$ with $\theta = 1/n$. We designed the non uniform sub-sampling distribution using (\ref{eq_opt_p}) with the true covariance matrix. We report relative errors in Frobenius norm as a function  of $T/n$ in Figure \ref{fig_fixed}. Each point in the plot is an average  over $50$ independent trials. We observe that when $m = 0.75n$ the non uniform sampling distribution matches closely the performance of the estimator with full data. It is clear that when $m = 0.50n$ and $m = 0.25n$ performance decreases (for uniform and non uniform sampling) as $m$ decreases, and non uniform sampling always outperforms uniform sampling. 
In Figure \ref{fig_adaptive} we evaluate our active method from Algorithm \ref{alg1} with batch size $B = 300$. We consider the same scenario as in Figure \ref{fig_fixed} for $m = 0.5n$ and $m=0.25n$. The proposed active covariance estimation method quickly learns the optimal sub-sampling distribution,  is always better than uniform sampling,  and matches the performance of the non uniform sub-sampling method obtained from the true covariance matrix.  
% * <antonio.ortega@sipi.usc.edu> 2017-10-28T07:36:38.705Z:
% 
% Reading below it was not clear whether this was synthetic data, since you were allowing a choice for the covariance matrix. Please clarify. 
% 
% ^.
Finally we show in Figure \ref{fig_adaptive2} the same experiment shown in Figure \ref{fig_adaptive}, but now with covariance matrix with parameter $\theta = 10/n$, which changes the effective rank  from $\erank(\mathbf{\Sigma}) = 9.08$ with $\theta = 1/n$ to $\erank(\mathbf{\Sigma}) = 17.86$. We observe that  the problem becomes more difficult since the estimation errors are larger. There is no major difference between uniform, and non uniform sampling, thus the advantages of the proposed active covariance estimation method are limited. This might be due to various effects including,  relative magnitudes of diagonal entries of $\mathbf{\Sigma}$, sampling budget $m$,  effective rank, and probability update algorithm. Moreover, since the effective rank did not change much (compared with $n$),  this experiment suggests the effective rank does not quantify effectively the problem difficulty. Also,  a more precise method to update the probabilities $\mathbf{p}$ might help improving the performance of the active covariance estimation algorithm.
\section{Proof of Theorem 1}
\label{sec_proofs}
%
%
%\subsection{Proof of Theorem \ref{th_concistency_fro}}
%\label{proof_consistency_fro}
We first need the following  concentration bounds
\begin{lemma}\label{lemma_tails}
Under the same assumptions of Theorem \ref{th_concistency_fro}, for any $\nu>0$ we have 
\begin{align*}
\Prob( \vert \widehat{\Sigma}_{ij} - \Sigma_{ij} \vert > \nu) \leq  2 \exp \left\lbrace -c_1 T \min\left(\nu^2/h_{ij}^2,  \nu/h_{ij} \right) \right\rbrace, \\
\Prob( \vert \widehat{\Sigma}_{ii} - \Sigma_{ii} \vert > \nu) \leq  2 \exp \left\lbrace -c_2 T \min\left(\nu^2 /h_{ii}^2,  \nu/h_{ii} \right) \right\rbrace,
\end{align*}
for off-diagonal and diagonal entries respectively.
\begin{proof}
The error events for off-diagonal entries satisfy
\begin{align*}
\vert \widehat{\Sigma}_{ij} - \Sigma_{ij} \vert > \nu &\Leftrightarrow  \vert \sum_{k=1}^T (y_i^{(k)}y_j^{(k)} - p_i p_j \Sigma_{ij} ) \vert >  \nu T p_i p_j.%\\
%\vert \widehat{\Sigma}_{ii} - \Sigma_{ii} \vert > \nu &\Leftrightarrow  \vert \sum_{k=1}^T ((y_i^{(k)})^2 - p_i\Sigma_{ii}  ) \vert >  \nu T p_i,
\end{align*}
%for off-diagonal and diagonal elements respectively.
We apply  Bernstein's inequality \cite{vershynin2016high} for sums of independent zero mean sub-exponential random variables, which combined with  the bound from Lemma \ref{lemma_bern_subgauss} leads to the desired bound. The proof for diagonal terms follows almost the same procedure.
\end{proof}
\end{lemma}
We also need the following geometric result which we state without proof.
\begin{lemma}\label{lemma_ell1_bound}
Let $\mathcal{A} = \lbrace \mathbf{x} \in \mathbb{R}^n: \Vert \mathbf{x} \Vert_1 > \epsilon \rbrace$, and $\mathcal{B}_i = \lbrace \mathbf{x} \in \mathbb{R}^n: \vert x_i \vert >\alpha_i \epsilon \rbrace$,  then for all $\epsilon>0$, and  $\alpha_i \in (0,1]$ that satisfy $\sum_{i=1}^n \alpha_i =1$, we have that $\mathcal{A} \subset \bigcup_{i=1}^n \mathcal{B}_i$.
%\begin{proof}
% Pick any $\mathbf{x} \in \mathcal{A}$ and suppose it  is not in $\bigcup_{i=1}^n \mathcal{B}_i$, therefore it must be that $\vert x_i \vert \leq \alpha_i \epsilon$ for all $i$, which implies that $\Vert \mathbf{x} \Vert_1 \leq \epsilon$, thus contradicting the original assumption, hence  proving the Lemma.
%\end{proof}
\end{lemma}
The proof of Theorem \ref{th_concistency_fro} starts by  bounding the probability of the event $\Vert \hat{\mathbf{\Sigma}} - \mathbf{\Sigma}\Vert_q > \epsilon \Vert \mathbf{\Sigma} \Vert_q$.  Pick a set of $\alpha_{ij} \in (0,1]$ such that $\sum_{i,j}^n\alpha_{ij} =1$, and apply Lemma \ref{lemma_ell1_bound} and the union bound to get
\begin{align*}
&\Prob(\Vert \hat{\mathbf{\Sigma}} - \mathbf{\Sigma}\Vert_q > \epsilon \Vert \mathbf{\Sigma} \Vert_q) =\Prob(\sum_{i,j} \vert\hat{\Sigma}_{ij} - \Sigma_{ij}\vert^q > \epsilon^q \Vert \mathbf{\Sigma} \Vert^q_q) \\
&\leq \sum_{ij}\Prob(\vert\hat{\Sigma}_{ij} - \Sigma_{ij}\vert^q > \alpha_{ij}\epsilon^q \Vert \mathbf{\Sigma} \Vert^q_q)\\
&=\sum_{ij}\Prob(\vert\hat{\Sigma}_{ij} - \Sigma_{ij}\vert > \alpha_{ij}^{1/q}\epsilon \Vert \mathbf{\Sigma} \Vert_q)\\
&\leq \sum_{i=1}^n 2 \exp \left\lbrace -c_2 T \min\left(\frac{\alpha_{ii}^{2/q}\epsilon^2 \Vert \mathbf{\Sigma} \Vert^2_q}{h^2_{ii}} ,   \frac{\alpha_{ii}^{1/q}\epsilon \Vert \mathbf{\Sigma} \Vert_q}{h_{ii}} \right) \right\rbrace + \\
& \sum_{i\neq j=1}^n 2 \exp \left\lbrace -c_1 T \min\left(\frac{\alpha_{ij}^{2/q}\epsilon^2 \Vert \mathbf{\Sigma} \Vert^2_q}{h^2_{ij}} ,   \frac{\alpha_{ij}^{1/q}\epsilon \Vert \mathbf{\Sigma} \Vert_q}{h_{ij}} \right) \right\rbrace.
\end{align*}
The last inequality follows from  Lemma \ref{lemma_tails} with constants $c_1$, $c_2$  appropiately chosen so the inequalities hold for all pairs $i,j$. We can further simplify   by choosing $\alpha_{ij} = {h^q_{ij}}/{\Vert \mathbf{H} \Vert^q_q}$
\begin{align*}
&\Prob(\Vert \hat{\mathbf{\Sigma}} - \mathbf{\Sigma}\Vert_q > \epsilon \Vert \mathbf{\Sigma} \Vert_q) \\
&\leq 2n \exp \left\lbrace -c_2 T \min\left( \epsilon^2\frac{ \Vert \mathbf{\Sigma} \Vert^2_q}{\Vert \mathbf{H} \Vert^2_q} ,   \epsilon\frac{ \Vert \mathbf{\Sigma} \Vert_q}{\Vert \mathbf{H} \Vert_q} \right) \right\rbrace \\
&+ 2(n^2-n)\exp \left\lbrace -c_1 T \min\left( \epsilon^2\frac{ \Vert \mathbf{\Sigma} \Vert^2_q}{\Vert \mathbf{H} \Vert^2_q} ,   \epsilon\frac{ \Vert \mathbf{\Sigma} \Vert_q}{\Vert \mathbf{H} \Vert_q} \right) \right\rbrace \\
&\leq 2n^2 \exp \left\lbrace - \frac{T}{\gamma} \min\left( \epsilon^2\frac{ \Vert \mathbf{\Sigma} \Vert^2_q}{\Vert \mathbf{H} \Vert^2_q} ,   \epsilon\frac{ \Vert \mathbf{\Sigma} \Vert_q}{\Vert \mathbf{H} \Vert_q} \right) \right\rbrace, 
\end{align*}
where $1/\gamma = \min(c_1,c_2)$.
The proof  can be finished by equating to $2/\eta$,  solving for $\epsilon$, and doing some $\min/ \max$ manipulations.
To derive the bound from (\ref{eq_theo_effRank_bound}) we bound  the  entries of $\mathbf{H}$ obtaining
\begin{align}\label{eq_bound_hii}
h_{ii} &=  \frac{\sigma^2 \Sigma_{ii}}{p_i } \leq  \frac{\sigma^2 \Sigma_{ii}}{\hat{p}}, \\ \label{eq_bound_hij}
h_{ij} &\leq \frac{\sigma^2 \sqrt{\Sigma_{ii}\Sigma_{jj}}}{p_i p_j } \leq \frac{\sigma^2 \sqrt{\Sigma_{ii}\Sigma_{jj}}}{\hat{p}^2 }.
\end{align}
 Then,   applying (\ref{eq_bound_hii}) and (\ref{eq_bound_hij}) followed by triangle inequality of the $\ell_q$ norm we have
\begin{align*}
\Vert \mathbf{H} \Vert_q &\leq \frac{\sigma^2}{\hat{p}} \left[\left(1-\frac{1}{\hat{p}^q}\right)\sum_{i=1}^n \Sigma_{ii}^q  + \frac{1}{\hat{p}^q}\left(\sum_{i=1}^n \Sigma_{ii}^{q/2}\right)^2 \right]^{\frac{1}{q}} \\
&\leq  \frac{\sigma^2}{\hat{p}} \left[\left(\frac{1}{\hat{p}^q} -1\right)^{\frac{1}{q}}\Vert \diag(\mathbf{\Sigma}) \Vert_q  + \frac{1}{\hat{p}}\Vert \diag(\mathbf{\Sigma}) \Vert_{\frac{q}{2}} \right]\\
&\leq  \frac{\sigma^2\tr(\mathbf{\Sigma})}{\hat{p}^2} \left[\left(1-\hat{p}^q \right)^{\frac{1}{q}} + 1 \right] \leq \frac{2\sigma^2}{\hat{p}^2} \erank(\mathbf{\Sigma}) \Vert \mathbf{\Sigma}\Vert.
\end{align*}
The last step uses the fact that $\Vert \mathbf{a} \Vert_q \leq \Vert \mathbf{a}\Vert_1$ for all $q \geq 1$, and the definition of effective rank.
\section{Conclusion}
\label{sec_conclusion}
We studied  covariance matrix estimation when the variables are  sub-sampled by a product with  Bernoulli $0-1$ variables. Variations of this model have been traditionally considered  in the analysis of missing data. We study an unbiased estimator for the covariance matrix and derive a novel estimation error bound in entry-wise $\ell_q$ norm. Our bound illustrates the subtle relations between covariance matrix parameters and sub-sampling distribution. Using this bound, we propose an active covariance matrix estimation algorithm that also produces an unbiased  estimator. We show with numerical experiments that the proposed active covariance estimation algorithm outperforms uniform sub-sampling, and closely matches non-uniform sub-sampling with complete knowledge of the true covariance matrix. 
\bibliographystyle{IEEEbib}
\bibliography{references}
\end{document}